\documentclass[nonacm, authorversion, screen]{acmart}

\usepackage{xcolor}

\begin{document}
\title{Quantitative AI Risk Assessments: Opportunities and Challenges}
\titlenote{This is a draft of a forthcoming article to appear in 
the \emph{Seton Hall Journal of Legislation and Public Policy.}
}

\author{David Piorkowski}

\email{djp@ibm.com}
\author{Michael Hind}
\email{hindm@us.ibm.com}
\author{John Richards}
\email{ajtr@us.ibm.com}
\affiliation{%
    \institution{IBM Research}
    \city{Yorktown Heights}
    \state{New York}
    \country{USA}
}

\begin{abstract}
Although AI systems
are increasingly being leveraged to provide value to organizations, individuals, and society, 
significant attendant risks have been identified~\cite{nist-rmf-2023,risk-atlas,IBM-ethics-board-POV-2024,owasp,mitre}
and have manifested~~\cite{propublica,Selbst2017,amazon-recruiting-2018,UK-exam-2020,zillow-2021,tesla-2019,imitation-attacks-2020,urlnet-2018}.
These risks have led to proposed regulations, litigation, and general societal concerns.

As with any promising technology, organizations want to benefit from the positive capabilities of AI technology while reducing the risks. The best way to reduce risks is to implement comprehensive AI lifecycle governance where policies and procedures are described and enforced during the design, development, deployment, and monitoring of an AI system. 
Although support for comprehensive governance is beginning to emerge~\cite{factsheets-product}, organizations often need to identify the risks of deploying an already-built model
without knowledge of how it was constructed or access to its original developers.
Such an assessment will quantitatively assess the risks of an existing model in a manner analogous to how a home inspector might assess the risks of an already-built home or a physician might assess overall patient health based on a battery of tests.

Several AI risks can be quantified using metrics from the technical community. However, there are numerous issues in deciding how these metrics can be leveraged to create a quantitative AI risk assessment.
This paper explores these issues, focusing on the opportunities, challenges, and potential impacts of such an approach, and discussing how it might influence AI regulations.
\end{abstract}

\maketitle


\section{Introduction}
AI systems\footnote{We define an AI system as 
\emph{"an engineered or machine-based system that
can, for a given set of objectives, generate outputs such as predictions, recommendations, or decisions influencing real or virtual environments~\cite{nist-rmf-2023}."}
}
are increasingly being used by 
commercial and noncommercial organizations
to provide new capabilities or to perform existing processes 
more effectively, or both.  As the adoption of AI  expands to high-stakes uses, combined with the fact that 
AI often exhibits higher variability than conventionally programmed systems,
concerns about the risk of AI
have increased. Several visible examples have dramatically illustrated these risks~\cite{propublica,Selbst2017,amazon-recruiting-2018,UK-exam-2020,zillow-2021,tesla-2019,imitation-attacks-2020,urlnet-2018} and risk taxonomies have emerged~\cite{nist-rmf-2023,risk-atlas,IBM-ethics-board-POV-2024,owasp,mitre}.
In addition to fundamental societal harm, 
these  risks can result in
negative brand reputation, customer/supplier/employee lawsuits, and increased regulatory scrutiny.
Organizations deploying AI are increasingly looking for ways to assess and mitigate these risks.

The creation of an AI system often occurs within a complex multistage lifecycle that begins with design and requirements, followed by model development and prompt engineering, model validation or quality assurance, deployment, and post-deployment monitoring.  One suitable approach to reduce risk is to instrument this lifecycle to collect and govern relevant ``facts"~\cite{factsheets-2019} about the process to ensure they comply with regulatory and organization policies intended to mitigate specific risks and prevent societal harm.
To support regulatory compliance requirements, we are seeing the emergence of  
systems to collect and manage these facts, 
enabling
transparency and governance~\cite{factsheets-product}, and expect mature organizations to leverage this technology.

However, organizations have a desire to assess the risks of an already-developed model, without knowing how the model was constructed or having access to its developers.  This can occur because mature AI transparency and governance may not yet exist in an organization, or because the model was obtained from a vendor,
an acquisition, or open source
that did not supply the necessary information. In situations like these, an organization may find it cost prohibitive or even impossible to recreate and collect the relevant facts about the model's development process. The organization can assess risk, based only on observing and evaluating the existing model's behavior. 

This requirement to assess risk by observing or evaluating something that is already built is quite common in other situations. A home inspector is hired to assess the current state of a house without knowing many details of how it was built.
A car inspection technician is often asked to assess the health of a car without knowing its repair history.  In each case, an expert uses their knowledge and various diagnostic tools to assess the entity ``as is", produce a report, and provide recommendations for improvement.
Such ``as is" assessments can be used to augment existing AI governance processes.
Given the success of quantitative assessments in other domains,
it is reasonable to explore the applicability of the concept to AI systems.

Since AI risks, particularly for ML models, can be quantified using metrics from the technical community~\cite{aif360,fairlearn,aix360,aip360,art360,uq360,ai-verify-foundation}, it is appropriate to consider how 
these metrics can be leveraged to create a quantitative AI risk assessment.
This paper will focus on the possibilities, difficulties, and potential impacts of a quantitative AI model risk assessment. 
Section~\ref{sec-regs} summarizes the approach taken by emerging AI regulations 
that will shape much of the risk landscape going forward. 
Section~\ref{sec-risk-assess}
discusses how a quantitative assessment can complement existing regulatory approaches.
Section~\ref{sec-dimensions} provides more details on the various dimensions of risk that a quantitative risk assessment might include.
Section~\ref{sec-ideal} describes desirable properties of metrics that form the basis for a quantitative assessment.
Section~\ref{sec-quant} broadens this discussion to consider desirable properties of a full assessment.
Section~\ref{sec-thoughts} discusses opportunities, challenges, and open research questions with a quantitative risk assessment.
Section~\ref{sec-conc} concludes the paper.

Most proposed AI regulation is qualitative. This is consistent with the lack of consensus on what a quantitative assessment should include~\cite{nist-rmf-2023}. One goal of this paper is to explore how a quantitative assessment could be added to existing AI system assessments, thereby providing a more holistic view of AI system risks.

\section{AI Regulations}
\label{sec-regs}
The area of AI regulation is dynamic and multifaceted. Various approaches and  proposals have been offered at both the national~\cite{EU-2021,UK-transparency-proposal-2021,us-alg-act-2022} and local~\cite{nyc-2021,illinois} levels. We expect regulatory activity will accelerate in the coming years with regulators refining their approaches based on feedback from the public, commercial vendors, technical organizations, legislative and standards bodies, and emerging case law. 

In mature technology segments, regulations and standards stipulate both process-oriented and measurement-oriented aspects. For example, the maximum allowable energy consumption in kilowatt hours per year for refrigerators and freezers of various capacities is specified in US Code \cite{USCode}. These limits are a small but critical part of a complex ecosystem of environmental and safety standards, labeling and documentation requirements, and testing procedures. This ecosystem has reached its present state over decades but continues to evolve subject to various well-specified processes of rule proposal, review, and codification.

Although current frameworks~\cite{nist-rmf-2023,eu-ai-act2024} for AI risk management have different emphases and objectives, they are primarily process-oriented and qualitative\footnote{Although primarily qualitative, regulations do not preclude quantitative measurement. However, most regulations do not require specific quantitative evaluations for AI systems and more importantly, have not yet specified what values measures should target.} in nature, stipulating that various processes should be followed and documented in the construction of an AI system.

Some industries are starting to use market forces to manage qualitative risks in AI systems. For example, the Data \& Trust Alliance is a consortium of companies that require HR vendors to provide certain forms of transparency about their AI processes through answers to a common set of several dozen questions~\cite{data-trust-2022}.

A few government bodies are venturing into the process of standardizing a few AI metrics.
Regulators in Singapore initially launched a pilot effort~\cite{ai-verify-2022} to understand  quantitative risk measurements of deployed commercial models with the goal of using this knowledge to inform future quantitative regulations.
This pilot led to formation of the 
AI Verify Foundation~\cite{ai-verify-foundation}, 
an open source foundation with over 150 members.
The foundation provides a framework and toolkit for exploring AI risk testing by the community. 
New York City enacted one of the first laws regulating the use of AI in hiring and promotion decisions \cite{nyc-2021}. 
It defines two impact ratios to compute potential bias with respect to multiple (intersectional) categories of people.
This assessment is to be periodically checked by independent external auditors (whose qualifications are unspecified) and publicly posted but no specific thresholds for compliance are specified.

Even though AI regulations and standards are in an early state, a more concrete operationalization can be envisioned and perhaps accelerated by a consideration of what AI risks can be
meaningfully measured. In what follows, we hope to provide a useful perspective on various measurable aspects of AI risk that could complement qualitative risk assessments.

\section{AI Risk Assessments} \label{sec-risk-assess}
The concept of an assessment or an audit of a system (or person or device) is not new to AI. It is common in any business process, as well as in everyday life, and is a key component of governance of that process.  Examples include home and car inspections, medical physicals, student and employee evaluations, and consumer product reviews.  In each case the entity is assessed based on some evaluation process and a report is produced.  The results from the report can provide a level of trust or confidence that the device/person is complying with some standards, 
or attaining some level of performance, enabling mitigating actions to be performed when it is not.
For example, if a student does not meet minimum testing standards, they may be required to repeat a course. If a patient does not achieve the expected value for an annual physical test, the physician may prescribe medication, therapy, or subsequent testing.

Independent of the evaluation criteria, more generally, assessments can be \textit{internal} or \textit{external}.
Internal assessments are performed by the organization that produced the entity being assessed~\cite{algo-audits-framework-2020}.
They have the advantage of being less costly and onerous than external assessments, but objectivity and conflicts of interests are concerns.
External or third party assessments utilize an outside organization to assess the entity.
Some external assessments are performed directly by government auditors,
others hire companies that specialize in assessments, often approved by regulators.  
For example, a company's source code may be audited by an external security auditing firm to validate that the company is following best software security practices.
External assessments provide greater objectivity, but can be more costly, both in terms of time and money, but also in terms of the risk of leaking proprietary information related to what is being assessed.

Internal and external assessments can be blended in various ways. For example, the US financial industry is regulated  under SR-11-7~\cite{sr-11-7}, which requires internal assessments for model risk management be performed by an independent group within the same financial institution. These assessments then feed into periodic external assessments/audits all with the goal of managing risk.  Significant repercussions exist for those firms that do not comply with the regulations.

Assessments can be \textit{qualitative}, \textit{quantitative}, or both. A qualitative assessment is one that is not meaningfully reduced to a set of numerical metrics. A description of how an AI model was developed is primarily qualitative (although certain aspects such as distributional properties of training data or accuracy measures with respect to test data are often measured). Qualitative assessment is generally broader and can, for example, provide a rich picture of a process, 
uncovering issues that could rarely be covered by a predefined suite of quantitative tests.
A quantitative assessment is one where well-defined metrics are computed as a way to assess one or more risks in an objective, standardized manner. Quantitative assessments tend to be more narrowly focused, 
often down to a specific concern,
but have the benefit of having specific values assigned to the measured phenomena.

Approaches that combine the qualitative and quantitative can provide a more holistic assessment.  For example, a physician utilizes both qualitative assessments (``How are you feeling today?  What is happening in your life?") and quantitative assessments (taking the patient's pulse, blood tests, x-rays) to assess a patient's health.  These assessments can be complementary in that the answers to qualitative questions can lead to further quantitative tests, and the results from quantitative tests can lead to additional qualitative questions.

\section{Quantitative Risk Dimensions} \label{sec-dimensions}
As the AI community moves toward more quantitative model risk assessments, we need to understand what risks can be meaningfully measured. In this section, we propose a non-exhaustive list of risk dimensions. Specifically, we focus in on risk dimensions that have metrics with certain desirable properties (to be discussed in Section~\ref{sec-ideal}) and that rely on nothing more than the model and data underpinning the AI system for their evaluation. For each, we describe the risk dimension, explain how the risk may manifest, and point to existing measurement approaches.

\subsection{Performance and Uncertainty}
A machine learning model that does not provide sufficiently accurate outputs
presents a fundamental risk if it is deployed.
Consider a model that predicts the price a customer is willing to pay for a particular service. If its predictions are too high, the business may lose customers. If its predictions are too low, the business may lose revenue.
Although considerable model development time is spent improving the accuracy of a model on known test data, it is difficult to assure that data received by the model after deployment continues to be accurately classified. 
For example, if a model is predicting credit worthiness of loan applicants, it is impossible to test all possible applicants, so there is a risk in how the model performs when it sees an applicant with characteristics quite different from those in its original training data.
There are many metrics to assess the accuracy of a model on a test dataset~\cite{100-page-ml-book}. 
More advanced techniques can look for variations in accuracy for different segments of the training data~\cite{frea}, can assess the uncertainty of a model's prediction~\cite{uq360}, or can generate ``challenging" synthetic test data {that intentionally differs from what the model was trained on~\cite{ai-testing}.
Other techniques can assess the quality of the training data used to create the model~\cite{dqai}.

In the case of generative models such as those that generate text, images, or video (e.g., ~\cite{mishra2024granite, achiam2023gpt, podell2023sdxl, xie2024sv4d}) there can often be more than one way to generate a ``correct'' response. Consider a simple case where a large language model (LLM) is asked to answer a question like, 
\begin{quote}
    \emph{Which country has the most people?}
\end{quote}
The complexity emerges from the abundance of possible correct and incorrect responses a model can generate. 
\begin{quote}
\emph{India} \\
\emph{As of July 1, 2023, the UN estimates that India has the most people, 1,438,069,596.} \\
\emph{India has 16 million more people than second place China.}
\end{quote}
are all correct, but not easily evaluated for correctness. 
Nevertheless, quantitative evaluation metrics exist to evaluate such models. For language models, a common approach is based on measuring how closely the model's output matches known correct or expected answers. This includes metrics such as BLEU~\cite{papineni2002bleu}, ROUGE~\cite{lin2004rouge}, and METEOR~\cite{banerjee2005meteor}.\footnote{METEOR goes a step further and accounts for synonyms.} More complex measures may try to discern if the meanings of the text are similar or more akin to human judgment. One such example is BERTScore~\cite{zhang2019bertscore}, which compares contextual embeddings instead of the text itself. These evaluations are inherently more prone to error because of the complexity of language. Yet, despite their limitations, they can still be informative about a language model's general accuracy. For models that generate images or video, evaluations still primarily rely on human evaluators.

Additionally, generative models can output factually inaccurate or untruthful content called \textit{hallucinations}. Research has shows that hallucinations are closely tied to a model's uncertainty and are more likely when the training data poorly represents the prompt~\cite{varshney2023stitch}. The problem of hallucinations has resulted in an active research community to quantify and mitigate this phenomenon~\cite{varshney2023stitch, rawte2023troubling, jesson2024estimating, ji2023towards, sanh2019distilbert}. More details including a recent taxonomy of hallucination mitigation strategies can be found in~\cite{tonmoy2024comprehensive}.

\subsection{Fairness}
Although machine learning, by its very nature, is a form of statistical discrimination, the discrimination becomes objectionable when it places certain privileged groups at systematic advantage over unprivileged groups. 
For example, it is reasonable to bias credit decisions towards applicants with higher salaries, but it is problematic to bias decisions towards a particular ethnic group.
Biases in training data, due to either prejudice in labels or under-/over-sampling, yields models with unwanted bias \cite{barocas-selbst}. A fairness assessment measures the likelihood that a model treats one group less favorably than another group even though the groups do not differ in a way that is relevant for the use case.
Systematically treating one group less favorably can be illegal, harmful to society, and result in litigation.
Thus, assessing the risk of such bias can help organizations avoid significant harm.
There are numerous metrics for measuring the fairness of a model and several tools (e.g., \cite{aif360,fairlearn,openscale}) have been developed.

In addition to making unfair decisions like traditional models, generative models' main feature of generating content gives rise to new forms of bias or propagation of existing societal biases in their output~\cite{mehrabi2021survey, bolukbasi2016man, hargittai2007whose}. 
Prior work has also shown how LLM output can be harmful, racist, classist, or stigmatizing~\cite{detectors-2024}.
Biases may also arise from the training data or input provided to models. For example, popularity bias arises when certain topics in the training data are overrepresented~\cite{dai2024bias}. Position bias refers to LLMs tendency to favor information in certain positions in the prompt~\cite{tang2023found}. 

Measuring fairness and bias in generated text is challenging due to the complexities in interpreting language. Current approaches rely on developing a (usually machine learning) classifier to detect the undesired behavior and score the model accordingly~\cite{detectors-2024}. Detectors based on traditional machine learning behave in a deterministic way and they can be used for quantitative measurement. Recent research~\cite{inan2023llamaguardllmbasedinputoutput,zeng2024shieldgemmagenerativeaicontent,han2024wildguardopenonestopmoderation,granite-guardian-3.0} uses another LLM as a ``judge'' to measure risks of the output of the first model.
Although promising, these techniques' reliance on another LLM result in nondeterministic evaluation, making consistent, repeatable evaluation difficult.

\subsection{Privacy}
Many privacy regulations (e.g., the European Union's General Data Protection Regulation \cite{gdpr}), mandate that organizations abide by certain privacy principles when processing personal information. 
Why is this relevant to machine learning? 
Studies have shown that a malicious third party with access to a trained ML model, even without access to the training data itself, can still infer sensitive, personal information about the people whose data was used to train the model.~\cite{shokri, fredrikson2015model}
For generative models, this includes the data used to fine-tune the model or provided as part of the prompt.
It is therefore crucial to be able to recognize and protect AI models that may contain, and thus potentially leak, personal information.
Research has expanded rapidly in this area. A summary of attacks, their associated metrics, and various mitigation approaches can be found in various toolkits~\cite{aip360,privml,privacy-meter,counterfit}. 
This can help in determining how to assess privacy risks for individual models based on distinctions such as black box vs. white box access and the type of model itself.

\subsection{Adversarial Robustness}
In addition to the risks of exposing private information mentioned above, attacks can also threaten to perturb the output of an AI system in ways advantageous to an attacker. Examples of this include the nearly invisible modification of input images to generate wildly erroneous classifications \cite{szegedy}. Model theft is also possible if the attacker can obtain output labels for chosen inputs~\cite{tramer-model-stealing-2016}.
If a model is not robust, a malicious actor could manipulate the model inputs to change its outcomes. With LLMs these attacks can expand to prompt attacks such as \emph{prompt injection} or \emph{jailbreaking}~\cite{rawat2024attackatlaspractitionersperspective,cornacchia2024mojemixturejailbreakexperts} that try to cause an LLM to provide information its creators did not intend.
Reducing the risks of adversarial attacks is a key consideration for models in the open, and often important for models accessible only within controlled intranets.
Researchers have provided various metrics, attacks, and mitigation techniques in several toolkits~\cite{art360,privml,privacy-meter,counterfit,cornacchia2024mojemixturejailbreakexperts}.

\subsection{Explainability}
Explanations can provide information to a human about why a model created a   \textit{particular} output. 
With generative AI, explanability can help a user 
understand which parts of the input the model relied on to produce an output~\cite{Chaudhury2022XFACTORAC,Murugesan2023,paes2024multilevelexplanationsgenerativelanguage}.
This complements the more global information associated with overall model transparency or understandability.
Explanations can reduce the risk of an inappropriate AI decision because humans can reason about and take corrective actions that lie outside the model's competence (e.g., deciding that someone near a decision boundary should be classified differently for reasons not considered by the model). Additionally, regulations such as \cite{gdpr,illinois}
require certain kinds of decisions to be explainable.
Although AI Explainability is a flourishing research topic~\cite{kush-book} with many tools available~\cite{aix360,lime,shap},
there is still a need for ways to assess the inherent explainability of a model.

\subsection{Value Alignment}
The ability of generative models to create varied content can make it difficult to guarantee that a model's output aligns with its creators' intent. The vast amounts of training data consumed to make these models effective can contain undesirable content that can be reflected in output. Examples include hateful and discriminatory language, violating norms, and deceptive language~\cite{detectors-2024}. Consequently, generative model output should be screened before being presented to an end user. Approaches include simple ones such as matching against a list of undesirable words, or more complicated machine-learning-based ones to detect more challenging content~\cite{elsherief2021latent, tillmann2023muted, webster2018mind, forbes2020social, mei2022mitigating}.

\section{Desirable Properties for Quantitative Assessment Metrics} \label{sec-ideal}
The basis for any quantitative risk assessment is a set of metrics 
that sufficiently capture various risk factors of a model.
There are two categories of metrics: \emph{individual metrics} that measure a particular risk and \emph{summary metrics} that combine several risk metrics. An individual metric, such as \emph{disparate impact} for assessing fairness, reflects a single aspect of model fairness. Summary metrics combine individual metrics to provide a more complete or more easily consumable picture. A summary metric could, for example, combine multiple fairness metrics to give an overall fairness score for the dimension of fairness. A summary metric could also combine multiple summary metrics (say accuracy, fairness, and adversarial robustness) into an overall model quality score that could be used to help in choosing a particular model from a set of available models.

These notions are frequently encountered elsewhere. For example, a student may take several exams throughout a course, but those individual exam scores will then be combined to provide an overall course grade. Course grades will then be combined for or an overall grade point average across all courses. 
Likewise, a product review may include many individual measurements which are then combined into an overall summary score allowing easier product comparison.
We see the same desire for AI risk assessments.

This section describes desirable properties for both kinds of metrics. Ideally, both individual and summary metrics will have these properties. Additional properties are also desirable for summary metrics due to the need for the consumer of the metric to understand how the individual metrics were combined.

\subsection{Individual Metrics}
Measurement theory provides guidance on the properties that make a metric more or less useful for assessing AI risk. Desirable properties for risk metrics include:

\begin{description}

\item [Deterministic:] \hfill \\ The metric value should be the same over repeated measurements if the 
output of the
evaluated model does not change.  Its value and interpretation should also not vary based on who is performing the measurement. 
More formally, the metric should operate as a mathematical function; for a given input, there is only one deterministic output.
If the metric entails some degree of randomness in its computation, care must be taken to differentiate changes to the metric value (due to this randomness) from real changes, 

\item[Valid:] \hfill \\ The metric should be measuring the construct we are actually interested in evaluating~\cite{cox2017statistical}.
For example, the number of "likes" for a chatbot response may not reflect response accuracy but may actually reflect the degree to which the chatbot is acting in a superficially pleasing manner.
Multiple measures, each with their own strengths and weaknesses, can sometimes be used to assess whether the underlying construct is being adequately captured.

\item [Monotonic:] \hfill \\ Changes in model behavior with respect to a risk should be consistently related to changes in the metric associated with that risk. For example, a monotonic relationship would be one in which the metric serving as a proxy for fairness improved as the true fairness of a model improved. A non-monotonic relationship would be one in which the metric sometimes improved and sometimes worsened as the true fairness of the model improved. Monotonicity makes the interpretation and comparisons of the metric easier~\cite{clapham2014concise}.

\item [Interval or ratio scale:] \hfill \\ Two models evaluated with the same metric but resulting in different values should indicate something meaningful about the difference in risk between the two models~\cite{pratt1995introduction}. 
If a metric has
an underlying interval scale, the difference between a score of 20 and score of 30 is the same as the difference between a score of 30 and a score of 40. 
If the metric has
an underlying ratio scale, it can be said that a score of 80 is twice as much as a score of 40.
These properties are desirable because they often match the intuition of non-experts using a metric.

\item [Applicable:] \hfill \\ The metric should be applicable across \textit{different} model types, thus enabling meaningful comparisons between them. For example, a metric that can be used to compare the outcomes of a decision tree and a neural network is more useful than a metric that can only be applied to one of these model types.

\end{description}

\subsection{Summary Metrics}

Similar to the individual metrics above, these summary metrics share the same desirable properties, but also have additional desirable properties related to the consumability of the summary metrics by others, as adapted from~\cite{norman1988psychology}.

\begin{description}
\item [Transparent:] \hfill \\ The summary metric should communicate how it was calculated and be able to explain how each of its metrics contributed to the final score. 

\item [Understandable:] \hfill \\ The summary metric should be understandable by the person expected to 
consume it. Depending on the technical expertise of the person, this may require additional content beyond the metric, such as explanations of the metrics, 
examples of
``good'' and ``bad'' values, and suggestions for improvement.
\item [Context-aware:] \hfill \\ 
Different use cases may need to summarize a collection of metrics differently.  For example, an employer may care only about a student's grades in certain classes or weight a particular course more than others.
Likewise, an AI risk owner may not be interested in the risk of adversarial attacks if they know the model will only be deployed in friendly environments.
Thus, 
the summary metric should consider such context in its calculations. If a use case increases the importance of one of the metrics being summarized, that importance should be reflected in the final summary.
\end{description}

Although certainly not a complete list, these properties are a starting point for evaluating some of the available metrics for AI risk evaluation today and understanding in what ways they may fall short.

\section{Additional Considerations for Quantitative AI Assessments} \label{sec-quant}

In addition to the properties of the metrics, there are additional considerations 
when applying one or more metrics for the purpose of assessing the risk of a particular AI model. In this section, we detail these additional considerations.

\subsection{Selecting the correct metric} 
\label{sec:correct-metric}
Determining how to measure a particular property is an important decision. In mature fields, metrics (such as glucose level, miles per gallon, blood alcohol level, or radon level) are well established and understood by their practitioners.  AI risk is a relatively new area, with new techniques and ideas for evaluating a model's risk dimensions being added regularly; 
a single risk dimension can have an overwhelming number of possible metrics.
For example, in the area of fairness, there are dozens of different metrics that not only focus on different views of what constitutes ``fairness'', but also have specific constraints on their applicability~\cite{aif360,kush-book}. Considerable assistance is needed to help the practitioner select, measure, and interpret the right metrics.

In addition to concerns about measuring the right thing, AI systems do not exist in a vacuum. The use of these systems may affect people. Consequently, there is an active research community that is tying societal concerns into the development and evaluation of AI systems. Some examples include investigating how to incorporate human values in building models~\cite{stray2020aligning, stray2021you}, identifying AI system evaluation gaps~\cite{hutchinson2022evaluation}, concretely mapping sources of harm caused by AI systems~\cite{suresh2021framework} and reducing barriers to AI system accountability~\cite{cooper2022accountability}. In addition to the AI-system-focused measures, a given use case should account for external impact and thus, may also need to consider metrics emerging from this work. 

\subsection{Interpreting a metric}
Closely related to the above is a metric's interpretability. A metric is an abstraction of an  associated risk. As with any abstraction, it can be useful but potentially misleading as it will never capture all aspects of the actual risk.
When interpreting a metric, a practitioner must endeavor to understand both what is and what is not being measured by it.
Metrics can be partial measures for the construct of interest.
Thus, the consumer of the metric should be aware that a metric is always a proxy for the characteristic of interest~\cite{thomas2022reliance}, and that its operationalization is based on a measurement model relying on multiple (often untested) assumptions~\cite{jacobs2021measurement}.

Interpreting multiple metrics carries additional risks. Thomas and Uminsky, proposed using multiple metrics to avoid gaming single metrics~\cite{thomas2022reliance}. However, selecting one appropriate risk metric, let alone multiple appropriate risk metrics for a given risk dimension, can be challenging. Even when done correctly, prior work on quantitative measures of AI systems has shown that these metrics are challenging to interpret without expertise~\cite{saha2020measuring, zhou2021evaluating}. 

Given this, the consumer of a metric 
should bear the responsibility 
to understand the strengths and weaknesses of the abstraction that is the metric.  Likewise, the metric provider should attempt to educate the consumer about what the metric measures and its appropriate use.  For example, high quality political polls come with a margin of error value. A consumer of such a poll that ignores these values in their analysis does so at their own risk. 

\subsection{Setting thresholds}
Once a metric is chosen, it can be useful to define acceptable values for that metric in the context of a specific use case. Sometimes, a threshold will be based on consensus within a field of practice and can be absolute. In internal medicine, for example, a blood sugar level less than 140 mg/dL (7.8 mmol/L) is categorized as normal. However, an acceptable value will often depend on the context.  For example, an acceptable limit for a car's speed will depend not just on the road but also on the present road conditions. Similarly, a model's threshold for acceptable performance may vary based on the business problem it is trying to solve. For example, a business may accept a model with a lower ability to identify possible fraud if the cost to the business of that fraud is relatively low. To take another example, a business may require a higher ability to detect toxic output from a chatbot if its conversations are directly with customers. 

Thresholds for individual metrics cannot generally be set in isolation as risk dimensions often trade off against one another. Research has shown that tuning a model to have a higher score on fairness will often lead to a lower score on predictive performance ~\cite{liu2022accuracy, wick2019unlocking, dutta2020there}. Other research has shown that model architectures yielding higher performance often make it more difficult to explain why a particular output was generated~\cite{baryannis2019predicting, bell2022s, gunning2019xai}. Business owners will generally have to consider tradeoffs between metrics across multiple properties of interest.

\subsection{Summarizing a dimension}
An assessment may contain several metrics in a single risk dimension, either to triangulate results, or to provide evaluations from different perspectives. It may be desirable to summarize a collection of a dimension's metrics into a single metric, similar to how a course grade summarizes the scores of all assignments and tests within the course. However, summarization presents its own set of technical and user difficulties. On the technical side, the question of how to combine metrics into a single score is nontrivial. Metric ranges, the relative importance of metrics, and how to account for conflicting results across metrics all contribute to the technical challenges. 
There is the danger that summarization obscures the nuances detailed by the individual metrics, or in the worst case, hides critical information. 

On the consumer/user side, familiarity with a given dimension, and how a user interprets the summary may also undermine the effectiveness of a summary. Additionally, two of the three desirable properties of summary metrics have user-centric elements: transparent and understandable. How a summary metric is calculated and how it is explained depends on the consuming user's knowledge and expertise. Any summary metric needs to be designed to consider who they are summarizing for to be effective.

\subsection{Summarizing across dimensions}
After choosing an appropriate metric or metrics for each risk dimension of concern and defining what are acceptable values for that metric or metrics, the next challenge becomes how to summarize the results for this collection of metrics into a higher level ``assessment score" suitable for a variety of roles (and technical expertise).  We see this strong desire for an overall score in areas such as education (a GPA), consumer reviews (scores out of 100), and energy efficiency ratings (ENERGY STAR or not).  
For an AI Risk Assessment, one can envision a score for each risk dimension: fairness, privacy, explainability, etc. and an overall risk score that combines each of these scores.  The challenge here is that different roles are likely to have different needs for this overall score and any associated descriptions of its meaning. Consider two roles, a government regulator and a model's business owner. The regulator will want to see how well the model's risk assessment is meeting the criteria for one or more regulations. In a summary, they may want to see the set of evaluations that show that the model meets the regulatory criteria. In contrast, a model's business owner may want to see how the various evaluations translate to additional revenue or other business-relevant key performance indicators. Across these and other cases, the way the information is summarized and contextualized can vary dramatically.

Although summary scores are quite popular and exist in many fields, there are risks to using them.  As mentioned earlier, any abstraction comes with the risk of it not representing 
key aspects of a property that its users need.  Since there can be a diversity of users of scores, it may be impossible to develop a summary that is useful for all.
Further, summaries often come with caveats on their use that often are not included in relaying the score to a larger population of non-experts. An example of this is political polls, which come with a confidence interval.  However, news organization often focus on the actual score without talking about the confidence interval, resulting in a public thinking that candidate A is ``beating'' candidate B, even though the difference in their scores is within the margin of error.
Thus, although we anticipate summary scores will be popular for AI Risk Assessments, we must take extra care to convey these caveats, particularly given the stakes of what is being assessed.

\section{Discussion} \label{sec-thoughts}
As the field increases its use of quantitative assessment, we see opportunities, challenges, and open research questions emerging. This section further explores some of these issues.

\subsection{The Interplay of Quantitative Measurements and Regulations} \label{sec-duality}
Regulations attempt to codify and enforce desirable properties of systems for the benefit of society.  
Organizations that are subject to these regulations desire precise language and detailed specifications to help them determine when their systems are compliant.
Having regulations specified in a readily actionable manner such as ``measure these metrics and ensure their values fall within these ranges" is desirable for both regulators and organizations striving to achieve compliance.
At present, however, what to measure, and what constitutes acceptable results, is still developing through a complex interplay of technological capabilities, emerging case law, and national culture.

It may be possible to accelerate this evolution.
As mentioned in Section~\ref{sec-regs}, based on their initial pilot~\cite{ai-verify-2022}, regulators in Singapore, 
with the support of 150 other members, launched the AI Verify Foundation~\cite{ai-verify-foundation} that provides an open source
toolkit for validating AI system performance in a standardized way that can help share knowledge in the space of quantitative assessments among technologists and regulators.

\subsection{The Value and Limitations of Measurement} \label{sec-abstraction}
Measurement is valuable if it produces useful insights.
But the act of measurement necessarily causes some information to be lost.
Consider the case of a credit score.
It may be a useful indicator of probability of loan repayment.
But it also neglects many other facts about individual loan applicants~\cite{humble-ai}.
Similarly, choosing a particular metric to characterize the fairness of a model decision reduces our sensitivity to other, possibly important, indicators of fairness.

This general problem is only exacerbated when multiple metrics, perhaps across multiple dimensions, are combined into a summary score. 
A summary score can be useful in deciding which model may be more suitable for an application.
But this should always be done with an awareness of what the summary score does not adequately capture.

Finally, for each dimension of risk it is important to consider both multiple measures and additional, inherently non-quantifiable information, to develop a more complete understanding of an AI system and its potential impact on individuals and society.

\subsection{The Tension Between Customization and Standardization}  \label{sec-customization-comparability}
There are two desirable properties of a quantitative risk assessment that seem to be in conflict. The first property is to have a consistent specification of the metrics to use and a clear strategy for aggregating these metrics into simple composite scores.  
This standardization would allow the direct comparison of two different AI systems, much as we do with consumer product reviews.
The second property is the need to customize an assessment based on the particular use case of the AI system.  
For example, assume our general palette of risk dimensions includes fairness and risk of adversarial attacks. If we are assessing these for a model predicting manufacturing defects, human fairness is not relevant.  Likewise, a model that is going to be deployed within an enterprise behind a firewall may be less concerned with the risk of adversarial attacks. 
Similar tensions between customization and comparison exist in other domains such as evaluating business performance~\cite{arya2005use} or health care delivery~\cite{sinsky2021standardization}. 
It remains an open question how to develop customized summary metrics that still support cross-model comparison.

\subsection{The Practicalities of a Quantitative Risk Assessment} \label{sec-practicalities}

To perform a quantitative assessment of any kind, one needs access to the entity being measured: home inspectors need to see the house, medical testing needs access to the patient, car inspections need to test the actual car. The form of access needed is dependent on the kind of test being run.  

For an AI system, access would minimally entail the ability to send inputs (ideally representative of the data the model will experience once deployed) and examine its outputs. This would not require access to the training data, the model development techniques or parameters, or the environment in which it was trained.  Providing more access (for example, to training data) could lead to more comprehensive measurements, just like a biopsy can measure more than visual inspection.

It may seem that the most minimal access is not difficult to attain. It does, however, require that a model's end point can be called, possibly from outside a firewall.
It also raises questions of access control and the potential exposure of sensitive data or proprietary information.
Alternatively, an assessment tool could be provided to the model's owners for their own use, an approach taken by AI Verify~\cite{ai-verify-2022,ai-verify-foundation}. But given the current state of the art, effective use of such a tool may require considerable training. Furthermore, direct access to the model may be impossible if it is purchased as a service from a model provider as is currently the case with LLM offerings.

Another key practicality is the cost of performing a quantitative assessment. Depending on the risk and assessment technique, this process can take significant time and computational resources.
As with traditional software testing, as the assessment effort increases, the likelihood that
the assessment will be a good representation of the risk increases.
This is particularly important in the space of LLMs where both the input and output domains are large and thus, challenging to assess thoroughly.

\subsection{Integration with Existing Risk Processes}
Some industries, such as finance or healthcare, have mature processes to assess, mitigate, and report on at least some forms of risk.
As more dimensions of AI risk emerge, these processes need to
adapt,
perhaps incorporating both new tools and new practices.
To maximize adoption, these tools and practices should be designed with an awareness of what is already in place.
Ideally, the technical implementation for new quantitative metrics will be pluggable to ease integration and avoid having to perform a wholesale replacement of existing processes.
The question of how quantitative metrics fit into existing risk assessment practices remains an open one.

\section{Conclusions} \label{sec-conc}
In this paper, we reflect on the possibility of incorporating more quantitative measurements into AI risk assessments. We report on the current state of AI risk assessment, which tends to be more qualitative than quantitative. We describe multiple risk dimensions, highlighting current quantitative measurement approaches.
We also posit 
desirable properties 
of quantitative metrics and their summaries.
Finally, we discuss challenges that need to be addressed as we move towards making quantitative assessments a reality. 

We are at an inflection point for assessing the risk of AI models. 
Hundreds of metrics for quantitative AI risk assessment are now available. The next step is to put them into practice and begin to address the challenges that we have identified.
In doing so, we might move towards more objective, consistent, and comparable evaluations of AI model risk.

\section*{Acknowledgements}
This work draws on the research of an extended IBM Research global team from Bangalore, Cambridge, Dublin, Haifa, and Yorktown Heights. 
We thank these researchers for their many contributions.
The authors also thank Abigail Goldsteen and the anonymous reviewers for feedback on earlier versions of this paper.

\bibliographystyle{IEEEtran}
\bibliography{IEEEabrv,main}

\end{document}